\documentclass[lettersize,journal]{IEEEtran}
\usepackage{amsmath,amsfonts,amssymb}
\usepackage{algorithmic}
\usepackage{algorithm}
\usepackage{array}
\usepackage[caption=false,font=normalsize,labelfont=sf,textfont=sf]{subfig}
\usepackage{textcomp}
\usepackage{stfloats}
\usepackage{url}
\usepackage{verbatim}
\usepackage{graphicx}
\usepackage{bm}
\usepackage{amsmath}
\usepackage{amsthm}
\newtheorem{lemma}{Lemma}

 \def\BibTexff{{\rm B\kern-.05em{\sc i\kern-.025em b}\kern-.08em
T\kern-.1667em\lower.7ex\hbox{E}\kern-.125emx}}

\begin{document}

\title{Over-the-Air Federated Learning in Cell-Free MIMO with Long-term Power Constraint}

\author{Yifan~Wang, Cheng~Zhang,~\IEEEmembership{Member,~IEEE}, Yuandon~Zhuang, Mingzeng ~Dai, Haiming~Wang, Yongming~ Huang,~\IEEEmembership{Senior Member,~IEEE}
\vspace{-20pt}
\thanks{Y. Wang, C. Zhang, and Y. Huang are with the National Mobile Communication Research Laboratory, Southeast University, Nanjing 210096, China, and also with the Purple Mountain Laboratories, Nanjing 211111, 
 China (e-mail: 220221104; zhangcheng\_seu; huangym@seu.edu.cn).\emph{ (Corresponding author: Cheng Zhang).}}
 \thanks{Y. Zhuang is with the Purple Mountain Laboratories, Nanjing 211111, 
 China (e-mail: zhuangyuandong@pmlabs.com.cn).}
  \thanks{M. Dai and H. Wang are with the Lenovo Research, Lenovo Group, Beijing 100085, China (e-mail: daimz4; wanghm14@lenovo.com).}}
 
\maketitle
\markboth{Journal of \LaTeX\ Class Files,~Vol.~14, No.~8, August~2021}%
{Shell \MakeLowercase{\textit{et al.}}: A Sample Article Using IEEEtran.cls for IEEE Journals}


\begin{abstract}
Wireless networks supporting artificial intelligence have gained significant attention, with Over-the-Air Federated Learning (OTA-FL) emerging as a key application due to its unique transmission and distributed computing characteristics. This paper derives error bounds for OTA-FL in a Cell-free MIMO system and formulates an optimization problem to minimize optimality gap via joint optimization of power control and beamforming. We introduce the MOP-LOFPC algorithm, which employs Lyapunov optimization to decouple long-term constraints across rounds while requiring only causal channel state information. Experimental results demonstrate that MOP-LOFPC achieves a better and more flexible trade-off between the model’s training loss and adherence to long-term power constraints compared to existing baselines.
\end{abstract}

\section{Introduction}
Cell-free multiple-input multiple-output (CF-MIMO) is anticipated to be a key enabling technology in next-generation wireless communication systems  \cite{8768014}. In CF-MIMO systems, multiple distributed access points (APs) within a given geographic area are connected to a central processing unit (CPU) via fronthaul links to enable coordinated transmission. This architecture provides enhanced spectral and energy efficiency, ultra-low latency, ultra-high reliability, and equitable quality of service (QoS) for user equipment (UEs).

With the rapid advancement of artificial intelligence (AI) technologies,  the development of communication networks that support distributed AI systems, such as Federated Learning (FL), has become essential \cite{9537935}. FL conducts training directly at the network edge using locally generated real-time data, reducing the computational load on central units and addressing local data privacy concerns \cite{9599369}. However, the effectiveness of FL is often limited by the communication efficiency between the central units and distributed nodes. In this context, Over-the-Air (OTA) computation has emerged as a promising technique for overcoming these limitations. By leveraging the superposition properties of analog signals, OTA-FL enables the simultaneous aggregation of distributed data across multiple devices, enhancing the efficiency of parameter transmission in FL systems and offering significant advantages over traditional methods  \cite{1}. Given the strengths of the CF-MIMO architecture and OTA computation in FL, exploring their coordination is essential for improving FL service capabilities.

Existing studies have focused on enhancing OTA-FL efficiency in traditional cellular systems, particularly in single-cell, multi-user scenarios, through the joint optimization of communication and computation resources. To design energy-efficient and practical communication systems, many of these studies incorporate energy and structural constraints. For example, \cite{10355909} explored the joint optimization of power harvesting systems and OTA-FL, proposing a transceiver design and device scheduling algorithm that demonstrates superior practicality and effectiveness. Additionally, \cite{9605599} addressed dynamic scheduling for power-limited devices, particularly under uneven local data distribution. Beyond communication resource optimization, \cite{12} investigated learning rate optimization for mean square error (MSE) reduction, introducing an enhanced Federated Averaging algorithm for local learning rate optimization and proposing a near-optimal beamforming solution.

In contrast to traditional cellular systems, OTA-FL within a CF-MIMO framework supports larger-scale distributed learning tasks, enhancing scalability and performance. The role of CF-MIMO in supporting FL was examined in \cite{11}, which proposed a practical implementation of OTA-FL integrated with CF-MIMO and conducted a performance analysis, highlighting the advantage of cell-free networks in reducing energy consumption compared to cellular systems for OTA-FL tasks. Additionally, the impact of wireless fronthaul errors between APs and the CPU was investigated in \cite{10622362}, where the authors developed transmit precoding and two-phase power allocation schemes based on sufficient statistics. Currently, research on OTA-FL in CF-MIMO systems mainly addresses communication errors, rather than focusing on the optimality gap \cite{10444710}, a metric more closely aligned with the AI model's loss function.

 Energy consumption has consistently been a critical issue in communication systems. Existing research on OTA-FL primarily focuses on short-term power constraints, with limited attention to long-term constraints. Comprehensive power constraints enable more efficient power allocation throughout the communication period by balancing immediate transmission needs with sustained power management. Furthermore, studies that do consider long-term constraints often operate under idealized channel state information (CSI) conditions.  An earlier study \cite{2} addressed long-term power constraints in OTA-FL systems and proposed a power allocation scheme. However, this study was limited to a single-antenna setup and required non-causal CSI, which is difficult to obtain accurately during each training round.   Lyapunov optimization offers an effective approach for addressing dynamic stochastic optimization problems, particularly in the context of wireless communications \cite{9036074}, making it well-suited for power allocation with long-term power constraint in OTA-FL systems.

\IEEEpubidadjcol 
In the context of CF-MIMO, there exists a research gap in optimizing OTA-FL performance under short-term and long-term power constraints, particularly in minimizing the optimality gap after several communication rounds. This paper studies this problem and  focuses on the joint optimization of power control and beamforming in a CF-MIMO scenario. The main contributions of this work are summarized as follows:
\begin{itemize}
    \item  We derive upper bounds of the first order and seond order of the OTA aggregation error and its based optimality gap expression. 
And we further formulate an optimization problem of joint beamforming and power control across multiple rounds with both the short-term and long-term power constraint.
    
        \item  An alternating iterative optimization algorithm, named MOP-LOFPC, is developed to separately tackle the beamforming vector at the AP side and the power control at the user side. In each round, beamforming is optimized to minimize the optimality gap component, while the power control across mutiple rounds with the long-term constraint are decoupled with the help of Lyapunov optimization, requiring only causal CSI.
    
    \item  Closed-form solutions are derived for both optimization variables, and their effectiveness is validated through extensive simulations. Compared to existing benchmarks, our proposed approach MOP-LOFPC  achieves a better and more flexible trade-off between the model’s training loss and adherence to long-term power constraints. 
\end{itemize}
\vspace{-5pt}
\section{System Model}
We consider an OTA-FL system within a cell-free network, which consists of 1 CPU, $L$ APs, and $K$ UEs, all working collaboratively to facilitate a FL task.\vspace{-5pt}
\subsection{Federted Learing Model}
 Assume that the FL model parameters are represented by 
$\mathbf{w}\in\mathbb{R}^{1 \times q}$, where $q$ is the size of the model parameters. A model training process involves identifying the optimal model $\mathbf{w}$ that minimises the global loss function $F(\mathbf{w})$. We define $T$ as 
 the maximum number of FL iteration rounds with $\mathcal{T}\triangleq\{1,\cdots,T\}$. A standard process of FL is as follows: 1) At each round $t \in \mathcal{T}$, the CPU broadcasts the global model of round $t$ to each UE. 2) The UE executes a stochastic gradient descent algorithm utilizing the local data set and the global model to derive the local model of round $t+1$. 3)  This local model is then uploaded to the AP and subsequently forwarded to the CPU via the fronthaul link. Finally, the CPU aggregates the all local models to obtain the global model of round $t+1$. This standard FL process is repeated until either the desired accuracy is achieved or the maximum number of training rounds is reached.

Let \(\mathcal{K}\) denote the set of UEs and \(\mathcal{D}_{k}\) represent the local dataset of UE \(k\), with the global dataset defined as the union of all local datasets, i.e., \(\mathcal{D}_{\mathrm{tot}} = \cup_{k \in \mathcal{K}} \mathcal{D}_k\). The loss function $f(\mathbf{w},\mathbf{x}_i,\tau_i)$ quantifies the error of model $\mathbf{w}$ on sample $\left( \mathbf{x}_i, \tau_i\right)$ , where 
 $\mathbf{x}_i$ and $\tau_i$ are the input and the label, respectively. The loss function of model $\mathbf{w}$ on dataset $\mathcal{D}_{k}$ is

\begin{equation}
F_k(\mathbf{w})=\frac1{D_k}\sum_{(\mathbf{x}_i,\tau_i)\in\mathcal{D}_k}f(\mathbf{w},\mathbf{x}_i,\tau_i).
\label{local function}
\end{equation}
We assume that each user is allocated a dataset of identical size, denoted as \( D \triangleq D_k = |\mathcal{D}_k|, \forall k \in \mathcal{K} \). Consequently, the global loss function of the model \(\mathbf{w}\) is given by
\begin{equation}
F(\mathbf{w})=\frac1{D_{\mathrm{tot}}}\sum_{k\in\mathcal{K}}D_kF_k(\mathbf{w})=\frac1K\sum_{k\in\mathcal{K}}F_k(\mathbf{w}),
\label{global function}
\end{equation}where \( 
D_{\mathrm{tot}} =|\mathcal{D}_{tot}|=\sum_{k\in\mathcal{K}}D_k\) denotes the size of global dataset. 
The local model of user $k$ in round $t+1$ is determined by the global model and local gradient in round $t$ \cite{13}, i.e., 
\begin{equation}
    \mathbf{w}^{t+1}_k=\mathbf{w}^{t}-\eta_t\nabla F_k\left(\mathbf{w}^{t
    }\right),
    \label{model}
\end{equation}where $\eta_t$ denotes the learning rate in round $t$. The ideal global model aggregated at the CPU should be $\overline{\mathbf{w}}^t=\frac{1}{K}\sum_{k=1}^{K}\mathbf{w}_{k}^t$.
 However, condisering the tranmission error in the model uploding process, the model aggregation error is non-negligible. The next section introduces OTA-FL in the cell-free scenario.
\subsection{Over-the-Air Communication for FL}
In this study, we assume that APs possess sufficient resources to ensure error-free downlink model broadcasting. To reduce uplink communication error in cell-free network, we employ the OTA computation technology. In contrast to conventional orthogonal multiple access technologies, all users transmit concurrently within the same time-frequency resource block during the communication process. We assume that the channel gain coefficient remains constant within each communication round but varies across different communication rounds.

In this study, we consider an uplink  communication system comprising  single-antenna UEs and  multiple-antenna APs, where the AP is connected to the CPU via a high-capacity backhaul link. Let $\mathbf{h}_{k,l}^t\in\mathbb{C}^{N_r\times 1}$ denote the complex channel gain vector from UE $k$ to AP $l$, where $N_r$ denotes the antenna number of AP.  
 The model information is aggregated at the AP $l$ using the following receive beamforming vector $\mathbf{r}_l^t\in\mathbb{C}^{N_r\times{1}}$, yielding the model parameter expressed as

\begin{equation}	\hat{\mathbf{w}}_l^t=(\mathbf{r}_l^{t})^{\rm{H}}\sum_{k=1}\nolimits^K\mathbf{h}_{k,l}^t\sqrt{p_k^t}\mathbf{w}_k^t+(\mathbf{r}_l^{t})^{\rm{H}}\mathbf{N}_l^t,
 \label{ap_signal}
\end{equation}where $p_k^t\in\mathbb{R}$ denotes the transmission power of UE $k$ at $t$-th round communication. $\mathbf{N}_l^t=\left[\mathbf{n}_{1,l}^{t},\cdots,\mathbf{n}_{q,l}^{t}\right]\in\mathbb{C}^{N_r\times{q}}$ represents the additive white Gaussian noise,  with each column vector being independently and identically distributed in a Gaussian distribution, i.e., $\mathbf{n}_{q,l}^t\sim{CN(0,\sigma^2\mathbf{I})}$ and $\sigma^2$ being the noise power.
The model information obtained at the CPU at $t$-th round communication is
\begin{equation}
\mathbf{z}^t=\sum_{l=1}^ L\hat{\mathbf{w}}_l^t=\sum_{l=1}^L(\mathbf{r}_l^{t})^{\rm{H}}\sum_{k=1}^{K}\mathbf{h}_{k,l}^{t}\sqrt{p_k}\mathbf{w}_{k}^{t}+\sum_{l=1}^L(\mathbf{r}_l^{t})^{\rm{H}}\mathbf{N}_l^t.
   \label{z}
\end{equation}

Recall the idea model, the aggregation error at $t$-th round communication can be expressed as
\begin{equation}
	\begin{aligned}
		&\bm{\varepsilon}^t=\mathbf{z}^t-\overline{\mathbf{w}}^t=\sum_{l=1}^L\sum_{k=1}^{K}m^t_{k,l}\mathbf{w}_{k}^{t}+\sum_{l=1}^L(\mathbf{r}_l^{t})^{\rm{H}}\mathbf{N}_l^t,
	\end{aligned}
 \label{error}
\end{equation}
where $m^t_{k,l}=(\mathbf{r}_l^{t})^{\rm{H}}\mathbf{h}_{k,l}^{t}\sqrt{p_k^t}-\frac{1}{LK}$. The error $\bm{\varepsilon}^t$ comprises two items, which are caused by channel fading and noise respectively. According to the (\ref{error}), it can be seen that the error caused by communication can be mitigated by adjusting  transmission power $\mathbf{p}^t=\left[p_1^t,\cdots,p_K^t\right]$ and receive beamforming vector $\mathbf{v}_t=\left[(\mathbf{r}_1^{t})^{\rm{H}},\cdots,(\mathbf{r}_L^{t})^{\rm{H}}\right]^{\rm{H}}\in\mathbb{C}^{{N_rL}\times{1}}$. To construct a green distributed  communication-computing integrated system, this study incorporates both instantaneous transmit power constraint and long-term transmit power constraint for each UE spanning the entire FL process, which be expressed as follows respectively:
 \begin{align}
     &\frac{1}{q}p_k^{t}\mathbb{E}\left(\left\|{\mathbf{w}_k^{t}}\right\|^2\right)\leq\hat{P}_k^{\max},\quad\forall k\in\mathcal{K}, \forall t\in\mathcal{T},
     \\&
     \frac{1}{qT}\sum_{t=1}^Tp_{k}^{t}\mathbb{E}\left(\left\|{\mathbf{w}_{k}^{t}}\right\|^{2}\right)\leq\hat{P}_{k}^{\mathrm{ave}},\quad\forall k\in\mathcal{K},
    \label{longcons}
\end{align}where $\hat{P}_k^{\max}$ denotes the maximum power budget for UE $k$ at each communication round, $\hat{P}_k^{\mathrm{ave}}$ denotes the long-term power budget for UE $k$  across all communication rounds.
\subsection{Problem Formulation}
According to a previous pioneering study \cite{2}, the gap between the loss function of the final round model $\mathbf{w}^T$ and the optimal loss function $F^{\star}$ can be quantified as 
\begin{equation}
    \begin{aligned}
&\mathbb{E}\left(F\left(\mathbf{w}^{T}\right)\right)-F^{\star} \\
&\leq\prod\limits_{t\in\mathcal{T}}M_{t}\left(F\left(\mathbf{w}^{0}\right)-F^{\star}\right)+\sum\limits_{t=1}^{T}J_{t}\left(\eta_{t-1}\Omega N+\eta_{t-1}^{2}\Omega^{2}W\right) \\
&+\sum_{t=1}^{T}\frac{J_{t}}{2}\left(\frac{1}{\eta_{t-1}}\left\|\mathbb{E}\left(\boldsymbol{\varepsilon}_{t}\right)\right\|^{2}+\left(S^{2}\eta_{t-1}\Omega+S\right)\mathbb{E}\left(\left\|\boldsymbol{\varepsilon}_{t}\right\|^{2}\right)\right),
\end{aligned}
\end{equation}where  $M_t = 1-(\Omega - 1)\mu\eta_t$, $J_t \triangleq 
\frac{\prod_{i=t}^TM_t}{M_t} $, $\Omega$ denotes the number of local model training epochs, $\mu$ characterizes the convergence rate of gradient descent, $N,W,S$ respectively captures the gradient variance bound, the gradient bounds, and the amplitude of gradient fluctuations. $\mathbb{E}\left(\boldsymbol{\varepsilon}_{t}\right)$ and $\mathbb{E}\left(\left\|\boldsymbol{\varepsilon}_{t}\right\|^{2}\right)$ denote the mean error and MSE in the $t$ round of model aggregation process.
 \begin{lemma} 
$\left\|\mathbb{E}\left[\boldsymbol{\varepsilon}_t\right]\right\|^2 $ and $\mathbb{E}\left(\left\|\boldsymbol{\varepsilon}_{t}\right\|^{2}\right)$ are bounded by (\ref{bias}) and (\ref{mse}), respectively, which can
be expressed as
\begin{align}
&\left\|\mathbb{E}\left[\boldsymbol{\varepsilon}_t\right]\right\|^2 
 \leq G^2\left\|\sum_{l=1}^L\sum_{k=1}^K m^t_{k,l}\right\|^2, 
\label{bias} \\
&\mathbb{E}\left[\left\|\boldsymbol{\varepsilon}_t\right\|^2\right] 
 \leq G^2\sum_{k=1}^K\left\|\sum_{l=1}^L m^t_{k,l}\right\|^2 + \gamma_t.
\label{mse}
\end{align}

Proof:
 By setting $\gamma_t=\sum_{l=1}^Lq\sigma^2(\mathbf{r}_l^{t})^{\rm{H}}\mathbf{r}_l^t$. Based on \cite{liu2020privacy}, we have the assumption about model bound $\mathbb{E}\left(\left\|\mathbf{w}_{k}\right\|^2\right)\leq G^2,\forall k\in\mathcal{K}$. Considering the irrelevance between different user models and the boundary of the models, the proof of Lemma 1 can be  completed. Details are omitted for page limit.
\label{lem-1}
\end{lemma}
By removing irrelevant terms, we can construct an equivalent optimization objective, such as
\begin{equation}
\begin{aligned}
        \Phi\left(\{p_{k}^t\},\{\mathbf{r}_{l}^t\}\right)
        &=\sum_{t=1}\nolimits^{T}\frac{J_{t}}{2}\left(\frac{\left\|\mathbb{E}\left(\bm{\varepsilon}_{t}\right)\right\|^{2}}{\eta_{t-1}}+B_t\mathbb{E}\left(\left\|\bm{\varepsilon}_{t}\right\|^{2}\right)\right)
        \\&\leq\sum\nolimits_{t=1}^{T}\phi[t],
\end{aligned}
\end{equation}where  $\phi[t]
        $$=A_t\left\|\sum_{l=1}^{L}\sum_{k=1}^{K}m^t_{k,l}\right\|^{2}$$+C_t\sum_{k=1}^{K}\left\|\sum_{l=1}^{L}m^t_{k,l}\right\|^{2}+B_t\gamma_t$ and $  A_t=\frac{J_{t}G^2}{2\eta_{t-1}}$,$ B_t=S^2\eta_{t-1}\Omega+S$, $C_t=B_tG^2$. The optimization problem can be formulated as
\begin{alignat}{1}(\mathbf{P}\mathbf{1}):
\min_{\{p_{k}^t\},\{\mathbf{r}_{l}^t\}} \quad & \frac{1}{T}\mathbb{E}\left\{\sum_{t=1}^{T}\phi[t]\right\},  \nonumber\\
\mbox{s.t.}\quad
&\frac{1}{T}\sum_{t=1}^{T}p_{t}^{k}\leq P_{k}^{\mathrm{ave}},\forall k\in\mathcal{K},  \label{constraint13} \\
&0\leq p_{k}^{t}\leq P_{k}^{\mathrm{max}},\forall k\in\mathcal{K}, \forall t\in\mathcal{T}\label{constraint14}, 
\end{alignat}where $P_k^{max}=\frac{q\hat{P}_k^{max}}{G^2},P_k^{ave}=\frac{q\hat{P}_k^{ave}}{G^2}, \forall k\in\mathcal{K}$. 
\section{Methodology}
The coupling of transmit power and receive beamforming in (P1) results in the problem being non-convex, which is challenging. The constraint  (\ref{constraint13}) further couples power variables across the time domain, adding complexity to the solution process.  To this end, we employ an alternating iterative optimization framework to decouple the optimization of power vector and beamforming vector. First, the receive beamforming vector is computed using a beamforming algorithm given the transmit power of all users. Subsequently, each user’s tranmsmit power is determined using a power optimization algorithm based on the receiving beamforming in the AP side. Causal CSI is considered in this paper for practical cases, which nonetheless makes the optimal solution to this problem hard to achieve due to the uncertainty of future channel states. 
\subsection{MOP Receive Beamforming}
 In this section, we propose the MOP receive beamforming algorithm for minimizing single-round component of optimality gap based on given power. Since the beamforming vectors are decoupled in different slots, their optimization can be addressed independently at each time slot. At each time slot, the optimization problem can be formulated as follows:
\begin{equation}
\begin{aligned}
    \operatorname*{min}_{\{\mathbf{r}_{l}^t\}} \hspace{0.5em} \phi[t]
&=A_t\left\|\mathbf{v}_{t}^{\rm{H}}\mathbf{u}^t-1\right\|^2+C_t\sum_{k=1}^K\left\|\mathbf{v}_{t}^{\rm{H}}\mathbf{u}^t_{k}-\frac{1}{K}\right\|^2\\&+B_t\gamma_t
\label{p3}
\end{aligned}
\end{equation}where 
$\mathbf{u}^t=[(\mathbf{h}_1^{t})^{\rm{H}},\cdots,(\mathbf{h}_L^{t})^{\rm{H}}]^{\rm{H}}\in\mathbb{C}^{{N_rL}\times1}$, $\mathbf{u}_{k}^t=[(\mathbf{h}_{k,1}^{t})^{\rm{H}}\hat{p}_k^t,\cdots,(\mathbf{h}_{k,L}^{t})^{\rm{H}}\hat{p}_k^t]^{\rm{H}}\in\mathbb{C}^{{N_rL}\times{1}}$, $ \mathbf{h}_{l}^{t}=\sum_{k=1}^K\mathbf{h}_{k,l}^{t}\hat{p}_k^{t}\in\mathbb{C}^{{N_r}\times1}$. By setting the first derivative of the objective function in
problem (\ref{p3}) to be zero, the optimal solution is calculated by
\begin{equation}
\begin{aligned}
\mathbf{v}_{t}^*=\frac{A_t\mathbf{u}^t+\frac{C_t}{K}\sum_{k=1}^K\mathbf{u}_k^t}{A_t\mathbf{u}^t(\mathbf{u}^{t})^{\rm{H}} +C_t\sum_{k=1}^K\mathbf{u}_{k}^t(\mathbf{u}_{k}^{t})^{\rm{H}}+B_tq\sigma^2\mathbf{I}}.\label{beamforming}
\end{aligned}
\end{equation}
\subsection{LOFPC Power Optimization}
To reduce the complexity of handling (\ref{constraint13}) within a finite time, we propose LOFPC  algorithm based on Lyapunov optimization to transform the power control problem from finite time to infinite time. By setting $H_k^t=\sum_{l=1}^L(\mathbf{r}_l^{t})^{\rm{H}}\mathbf{h}_{k,l}^{t}$, $\hat{p}_k^t=\sqrt{p_k^t}$, while eliminating irrelevant terms, a new optimization problem can be formulated as
\begin{alignat}{1}(\mathbf{P}\mathbf{2}):
\min_{\{p_{k}^t\}} \quad & \operatorname*{lim}_{T\to\infty}\frac{1}{T}\sum_{t=1}\nolimits^{T}\mathbb{E}\{A_t\|\sum_{k=1}^{K}\hat{p}_k^tH_{k}^{t}-1\|^{2}+\label{p2}
\nonumber\\&C_t\sum\nolimits_{k=1}^{K}\|\hat{p}_k^tH_{k}^{t}-\frac{1}{K}\|^{2}\},  \nonumber \\
\mbox{s.t.}\quad
&\operatorname*{lim}_{T\to\infty}\frac{1}{T}\sum\nolimits_{t=1}^{T}p_{k}^{t}\leq P_{k}^{\mathrm{ave}},\forall k\in\mathcal{K},  \\
&0\leq p_k^t\leq P_{k}^{\mathrm{max}},\forall k\in\mathcal{K},  t\in\mathcal{T}. 
\end{alignat}In order to decouple the temporal coupling in (P2), a virtual queue is established for each UE to represent the current cumulative energy consumption, which can be expressed as
\begin{equation}
    q_k[t]=max\{q_k[t-1]+(\hat{p}_k^t)^2-P_k^\text{ave},0\}, q_k[0]=0.
\label{q}
\end{equation}Next, we will outline the method for utilizing virtual queues to convert long-term power constraints into processes that can be managed within each individual time slot. According to (\ref{q}), we have the following result:
\begin{equation}
    \frac1T\sum_{t=1}^T\left((\hat{p}_k^t)^2-P_k^\mathrm{ave}\right)\leq\frac1T(q_k[T]-q_k[0])=\frac1Tq_k[T],
\end{equation}where the long-term power constraint can be met when $\lim_{\mathrm{T}\to\infty}\frac1Tq_k[T]=0$.
Based on the virtual queue, the Lyapunov drift function can be defined as
$
\Delta L_k(t) = \mathbb{E}\{\frac12\left(q_k[t]^2 - q_k[t-1]^2\right) \mid q_k[t-1]\}
$. The Lyapunov drift plus penalty function is defined as $\Delta L(t)+V\mathbb{E}\{\phi[t]\}$, where the penalty factor $V>0$ is introduced to balance the average power stability and the communication efficiency of FL. If \( V \) decreases, the power converges more rapidly to satisfy the average power constraint. As \( V \) increases, better optimization performance can be achieved.

According to (\ref{q}), we can drive that:
\begin{equation}
    \begin{aligned}
\Delta L_k(t)
\leq \frac{1}{2}\left((\hat{p}_k^t)^2-P_k^\mathrm{ave}\right)^2+q_k[t-1]\left((\hat{p}_k^t)^2-P_k^{\mathrm{ave}}\right),
\end{aligned}
\end{equation}Therefore, we can formulate a new optimization problem, which can be expressed as \begin{alignat}{1}
(\mathbf{P}\mathbf{3}):\min_{\{\hat{p}_k^t\} }\quad & \mathbb{E}\{\xi[t]+V\phi[t]\},  \nonumber\\
\mbox{s.t.}\quad
&0\leq\hat{p}_{k}^{t}\leq\sqrt{P_{k}^{\mathrm{max}}},\forall k\in\mathcal{K}, \forall t\in\mathcal{T},
\end{alignat}where $\xi[t]=\sum_{k=1}^{K}\frac{1}{2}\left((\hat{p}_k^t)^2-P_k^\mathrm{ave}\right)^2+q_{k}[t-1]\left((\hat{p}_k^t)^2-P_{k}^{\mathrm{ave}}\right)$. 
We can differentiate the optimization objective and find a closed-form solution for the optimization variable $\hat{p}_k^t$ according to Cardano formula, which can be written as 
\begin{equation}
\hat{p}_{k}^{t}=\begin{cases}\sqrt[3]{-\frac{b_k^t}{2} + \sqrt{D}} + \sqrt[3]{-\frac{b_k^t}{2} - \sqrt{D}},&D\geq0\\\\ r \cos\left(\frac{\theta + 2\pi n}{3}\right), \quad n = 0, 1, 2,&D<0,\end{cases}\label{power}
\end{equation}where $a_k^t = q_k[t-1] - P_k^\mathrm{ave} + V(A_t + C_t)\|H_k^t\|^2$, $b_k^t = V\left(A_t\Re(H_k^t \sum_{i \neq k}^{K} \hat{p}_i^t H_i^{t*}) - (\frac{C_t}{K} + A_t)\Re(H_k^t)\right)$, $	D = \left(\frac{b_k^t}{2}\right)^2 + \left(\frac{a_k^t}{3}\right)^3$, $r = 2 \sqrt{-\frac{a_k^t}{3}},\theta = \cos^{-1}\left(-\frac{b_k^t}{2r}\right)$.
Based on the previous two sections, the alternating optimization algorithm for solving (P1) can be derived, as shown in Algorithm 1,  denoted as MOP-LOFPC.
\begin{algorithm}[!h]
{{  \caption{MOP-LOFPC.}
    \label{alg:AOA}
    \renewcommand{\algorithmicrequire}{\textbf{Input:}}
    \renewcommand{\algorithmicensure}{\textbf{Output:}}
    \begin{algorithmic}[1]
        \REQUIRE  $\mathbf{h}_{k,l}^t$, $\forall k\in\mathcal{K}$, $\forall l\in\mathcal{L}$, $\forall t\in\mathcal{T}$ 
        \ENSURE transmit power $\mathbf{p}^t$, beamforming vector $\mathbf{v}_t$, $\forall t\in\mathcal{T}$    
        \STATE \textbf{Initialize:} $t \gets 1$
        \WHILE{$t \leq T$}
             \STATE \textbf{Set:} Iteration counter \( i = 0 \),  $\mathbf{p}^t_i \gets P^{\text{ave}} \cdot \mathbf{1}_K$
    \REPEAT
        \STATE Increment iteration counter \( i = i + 1 \)
        \STATE \textbf{Step 1:} 
        Update the beamforming \( \mathbf{v}_t^{i} \) using (\ref{beamforming}), given \( \mathbf{p}^t_{i-1} \)  and $\mathbf{h}_{k,l}^t$, $\forall k\in\mathcal{K}$, $\forall l\in\mathcal{L}$;
        \STATE  \textbf{Step 2:} 
        Update the power \( \mathbf{p}^t_{i} \) using (\ref{power}), given $\mathbf{v}_t^{i}$  and $\mathbf{h}_{k,l}^t$, $\forall k\in\mathcal{K}$, $\forall l\in\mathcal{L}$;
    \UNTIL{convergence or maximum iterations reached}
       \STATE \textbf{Set:} $\mathbf{p}^t=\mathbf{p}^t_{i}$, $\mathbf{v}_t=\mathbf{v}_t^{i}$, $t=t+1$
        \ENDWHILE
    \end{algorithmic}}}
  
\end{algorithm}
\vspace{-12pt}
\section{Simulation Results}
\subsection{Simulation Setup}
We consider randomly placing 3 UEs and 6 APs in a  $500 m \times 500 m$ area. Each UE is equipped with a single antenna, while each AP is equipped with 4 antennas. Channel fading is obtained by multiplying large-scale fading and small-scale fading. We assume that the small-scale fading follows a Rayleigh fading model. The large-scale fading is characterized by Log-Distance Path Loss Model in an urban environment. The carrier frequency is set to 2.4 GHz, and the bandwidth is 20 MHz. All UEs have the equal power constraints, with an average power \( P_k^{\mathrm{ave}} = 0.3 \) W and a maximum instaneous power \( P_k^{\mathrm{max}} = 0.5 \) W. The noise power is \( -101 \) dBm.

We adopt the ridge regression task setup as described in \cite{2} with a fixed learning rate. The sample-wise loss function is given by $f(\mathbf{w},\mathbf{x},\tau)=\frac12\|\mathbf{x}^\mathrm{T}\mathbf{w}-\tau\|^2+\rho R(\mathbf{w})$, where the generated sample vector $\mathbf{x}\in\mathbb{R
}^{q}$ with $q=10$  and the
regularization function is $R(\mathbf{w}) = \|\mathbf{w}\|^2$ with the hyperparameter $\rho = 5\times 10^{-5}$. We consider the data samples are evenly distributed among
users with identical size $D_k = 1000,\forall k \in \mathcal{K}$. Its label is obtained by $y = x(2) + 3x(5) + 0.2z$, where $x(n)$ represents the n-th  element of $\mathbf{x}$, and $z$ denotes the observation noise with i.i.d. Gaussian distribution, i.e., $z \sim \mathcal{N}(0,1)$.
\subsection{Baselines}
Three  power control algorithms for comparison are given as follows.

\textbf{Fixed power strategy:} Each user employs a fixed transmission power in each communication round 
$p_k^{t}=P^\text{ave}_k,\forall k\in\mathcal{K}.$

\textbf{Ci power strategy:} In each time slot \( t \), each UE performs power optimization for \( \phi[t] \). Formally, similar to the channel inversion strategy, the transmit power is given by \( p_k^t = min \left\{\left(\frac{\Re(H_k^t)}{K\|H_k^t\|^2}\right)^2, P^{max}_k\right\}\).

\textbf{Lgr power strategy:} The Lagrangian dual method is used to solve the time coupling issue arising from long-term power constraints (\ref{constraint13}) in power control problem \cite{2}. However, this algorithm is an offline power control strategy and requires non-causal  CSI, i.e., the perfect CSI of all communication rounds. 

The proposed beamforming strategy is the MOP algorithm, while the MRC beamforming strategy serves as a baseline for comparison. This strategy is a reasonable maximum ratio combining algorithm for cell-free scenarios, as referenced in \cite{11}. The receive beamforming at each AP is a weighted sum of the channels from all UEs to the AP, with weights inversely proportional to the large-scale fading of the channels.

\begin{figure*}
	\setlength{\abovecaptionskip}{-5pt}
	\setlength{\belowcaptionskip}{-10pt}
	\centering
	
	\begin{minipage}[t]{0.33\linewidth}
		\centering
		\includegraphics[width=2.4in]{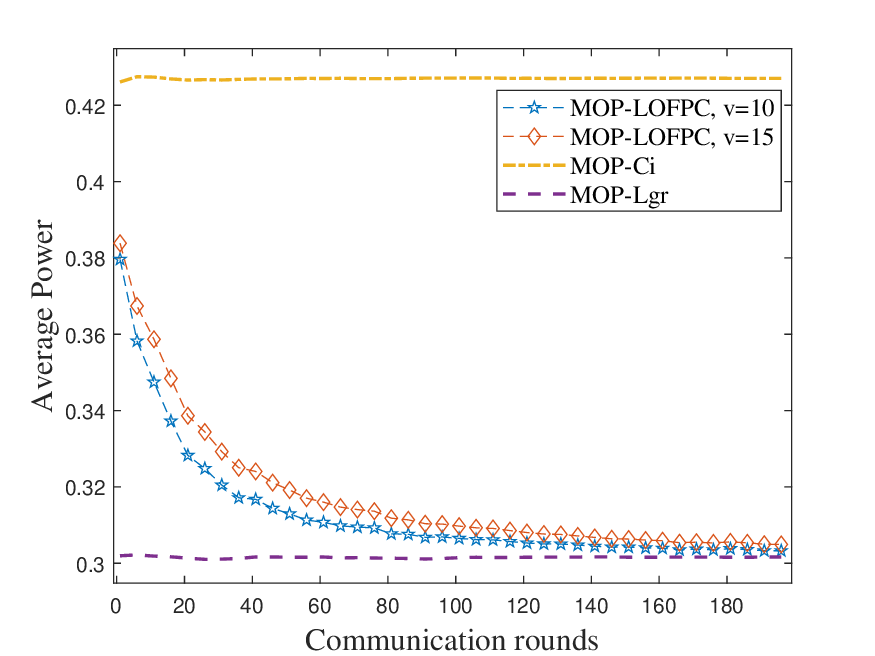}
		 \caption{\scriptsize Average power versus communication rounds.}
		\label{figure1}
	\end{minipage}
 \begin{minipage}[t]{0.33\linewidth}
		\centering
		\includegraphics[width=2.4in]{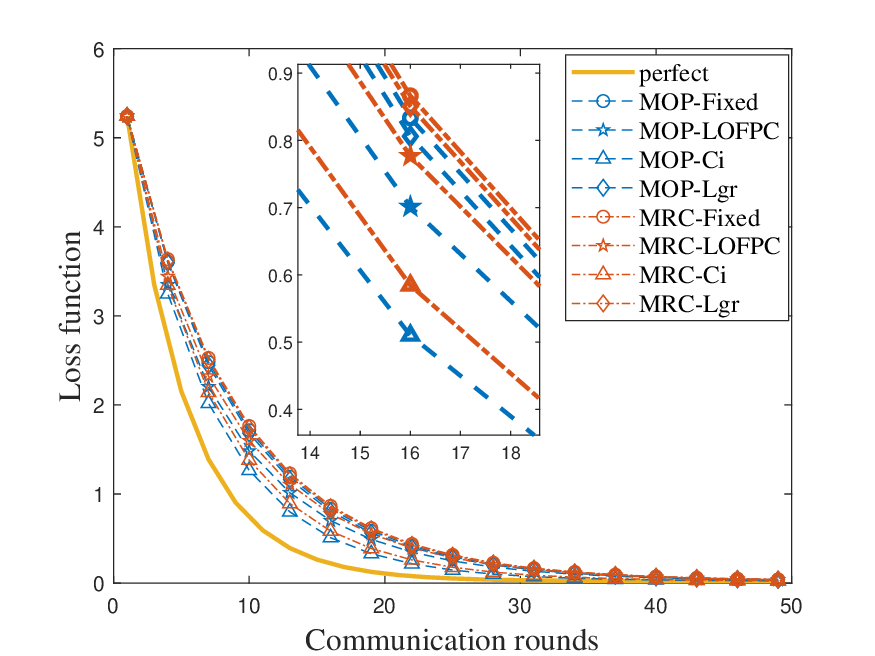}
		\caption{\scriptsize Loss function versus communication rounds.}
		\label{figure2}
	\end{minipage}%
	\begin{minipage}[t]{0.33\linewidth}
		\centering
		\includegraphics[width=2.4in]{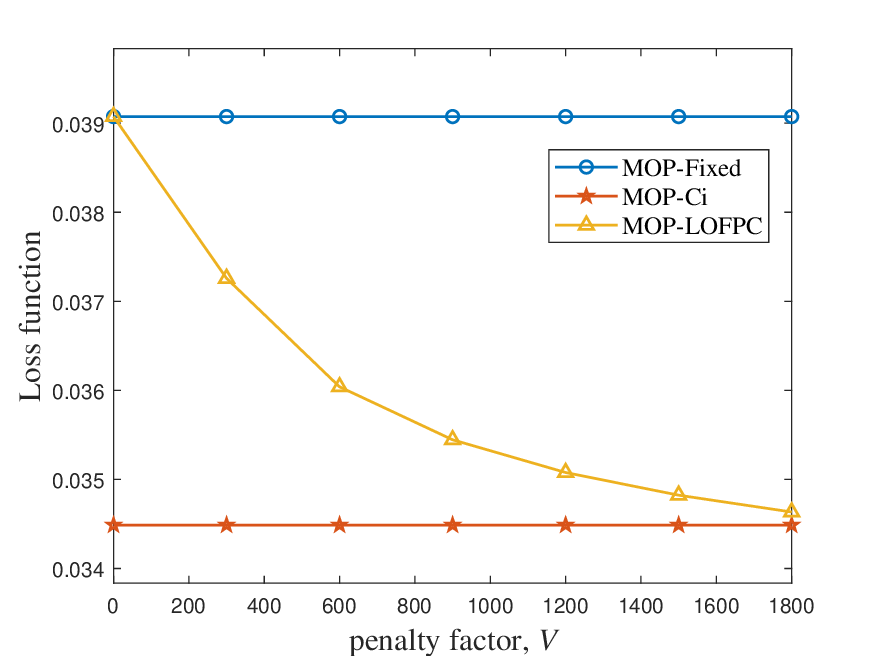}
		\caption{\scriptsize Loss function versus penalty parameter.}
		\label{figure3}
	\end{minipage}
\end{figure*}\vspace{-10pt}
\subsection{Results and Discussions}
Fig. 1 shows the satisfaction of long-term power constraints across communication rounds for different schemes. The MOP-Lgr algorithm consistently meets the long-term constraint, while the MOP-LOFPC algorithm converges after approximately 100 iterations. Conversely, the MOP-Ci algorithm fails to satisfy the constraint compared to the others.

Fig. 2 illustrates the learning loss across various power control and receive beamforming schemes. The   two LOFPC algorithms consistently outperform both the fixed and  Lgr algorithm under the same beamforming strategy. The Lgr algorithm enforces stricter long-term power constraints, while the fixed algorithm lacks optimization flexibility. Although the Ci method yields better performance than LOFPC and Lgr, it fails to satisfy long-term power constraints. Additionally, the MOP beamforming scheme consistently outperforms MRC scheme, as MOP directly optimizes the gap to the optimal loss function, whereas MRC focuses solely on channel gain and does not quantify the impact of communication errors on FL.

Fig. 3 shows that the MOP-LOFPC algorithm balances improved optimization objectives and accelerated convergence to the long-term power constraint by adjusting the penalty parameter \(V\). Recall that Fig. 1 and Fig. 3, we can see that a larger V can achieve better model training results, but it takes longer time for the average power to converge. A smaller V can achieve faster average power convergence, but the model training effect is worse under the same training rounds.
\vspace{-5pt}
\section{Conclusion}
This paper introduces a communication resource allocation algorithm to enhance energy efficiency in distributed AI systems within a cell-free framework. We formulated an optimization problem linking communication error to the optimality gap and proposed a joint power control and beamforming algorithm, named MOP-LOFPC,  based on the alternating iterative optimization framework. Experimental results demonstrate that the MOP-LOFPC provides a more effective and adaptable balance between minimizing the model's training loss and maintaining compliance with long-term power constraints, outperforming existing benchmarks.
\bibliographystyle{ieeetr}
\bibliography{IEEEabrv,ref.bib}
\end{document}